\documentclass[12pt]{article}
\usepackage{amsmath,amssymb,authblk,graphicx,times}

\title{The energy landscape of a simple \\ neural network}
\author[1,2]{Anthony Gamst}
\author[2]{Alden Walker}
\affil[1]{University of California, San Diego}
\affil[2]{Center for Communications Research, La Jolla}
\date{}               

\newcommand{\R}{\ensuremath{\mathbb{R}}}

\begin{document} \maketitle

\begin{abstract} 
We explore the energy landscape of a simple neural network. 
In particular, we expand upon previous work demonstrating 
that the empirical complexity of fitted neural networks 
is vastly less than a naive parameter count would suggest 
and that this implicit regularization is actually 
beneficial for generalization from fitted models.
\end{abstract}

\section{Introduction}

Neural networks have become very popular due to their ability to fit 
complicated high dimensional functions (e.g. determining the breed of 
dog featured in a photo). In many cases, the most successful networks 
are enormous, with many millions of parameters. Although typical training 
data sizes are also quite large, the amount of data is often miniscule when 
the dimension of the feature space is considered or when compared to the 
theoretical fitting capability of the network. We typically explain away 
the first issue by observing that the region of interest in a high dimensional 
feature space is likely to be much lower dimensional, so as long as we have a 
representative training sample, a fitted model should be accurate where it 
needs to be. However, we are left with the problem that the networks we use 
are often much too large, in the sense that they should be capable of 
completely reproducing the training data. In practice, this does not occur; 
in other words, this problem is not actually a problem, which leaves us with 
the question: why?

In~\cite{random,empirical}, we show that random neural networks 
(which includes standard initialized networks under training) are 
far less complex than they theoretically could be, and we show that 
trained neural networks retain this simplicity throughout training. 
In other words, the answer to the question of why networks do not 
overfit is:
\begin{enumerate}
  \item the complexity of a network increases smoothly with training time,
and
  \item there is an empirical upper bound on complexity which is much less 
than the theoretical limit.
\end{enumerate}
In this paper, we expand upon (2) and explain why it can be beneficial.
Specifically, we will construct a small neural
network which is, by design, capable of fitting, but essentially
incapable of learning, a given complex ``high frequency'' function.
We study the energy landscape of this network in an effort to
understand the distribution and size of the local minima and the general
topography of the weight space. We find that local minima are common;
the weight space is rough, with many long, flat valleys; and that the
domains of attraction of the global minima are very narrow.
The global minima in the weight space correspond to a perfect
fit of the data, but the local minima correspond to a ``de-noised''
version of the function which has had the high-frequency components
removed.  That is, the energy landscape has a regularizing effect,
and it is essentially impossible to reach a global minimum while
training.

The definition of ``noise'' is problem dependent: we might want to
actually fit the high-frequency components in our example below,
and we can do that by increasing the size of the network.  Our point
is not that networks in general cannot learn the function we construct
but that networks large enough to fit the function perfectly
might still not be large enough to actually learn it in practice, and
when they cannot learn it, they do a good job of learning just the
low frequency components.

The paper~\cite{ballard}, discovered after the initial draft of this
paper was written, also explores the energy landscape of neural networks,
although their theme is different from ours.
We remark on one point in the conclusion of~\cite{ballard} which appears
at odds with our assertion that it can be difficult to find a global minimum
or a local minimum which is competitive with it: they observe the opposite;
they find it quite easy to find a global minimum.  In fact, both observations 
can be reconciled: Given a trained network, which our previous work has shown
is likely to produce smooth (low complexity) output, it should be relatively 
easy to train a second network to match the output of the first; good local
minima for the two networks are likely to be similar. (And randomly-initialized
networks tend to produce smooth, low complexity output.) On the other hand, a
network constructed (not trained) to produce a specific non-smooth
(high complexity) output would be difficult for another network
(with the same architecture) to train to.

\section{Construction of a perfect fit}

Our goal is to exhibit a function and a network structure such that
the network can perfectly fit the function but never does under
training.  Note that the only way to produce such a function is
as the output of the network itself, because otherwise, we could
never find the perfect fit.  For simplicity, we construct a network
with one input and one output, and we have $K$ hidden densely connected
layers of $N$ nodes each.  We call this a $K \times N$ network.
The hidden layers all have relu activation,
and the output layer has a linear activation.  We will carefully
construct the weights of the network in two blocks.  Choose $N_1$ and
$N_2$ such that $N_1 + N_2 = N$.  The first
block, arbitrarily on the left side of the network, uses $N_1$ nodes
in each hidden layer to produce a high-frequency sawtooth wave
in a manner similar to \cite{upper_bound}.  The second block, on the
right, uses $N_2$ nodes in each hidden layer to produce a relatively
simple low frequency spline.  In fact, we only actually use the
first layer on the right: the subsequent layers are the (relued)
identity function.  The output node simply takes the sum of the 
high frequency sawtooth and the low frequency spline. We refer to 
this as a $K\times (N_1+N_2)$ network.

\begin{figure}[ht]
\begin{center}
\includegraphics{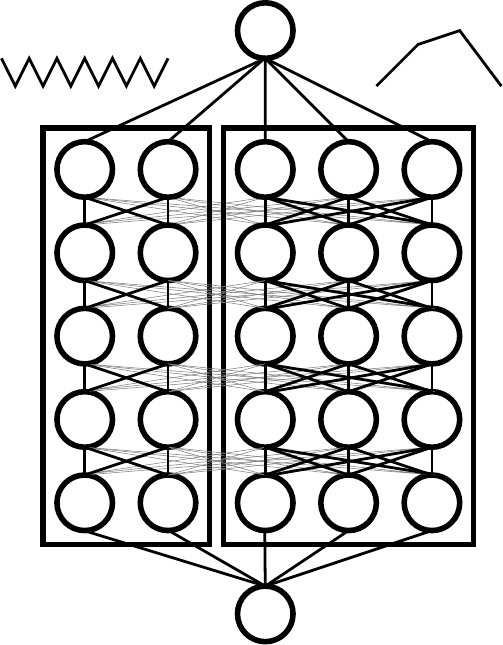}\includegraphics[scale=0.45]{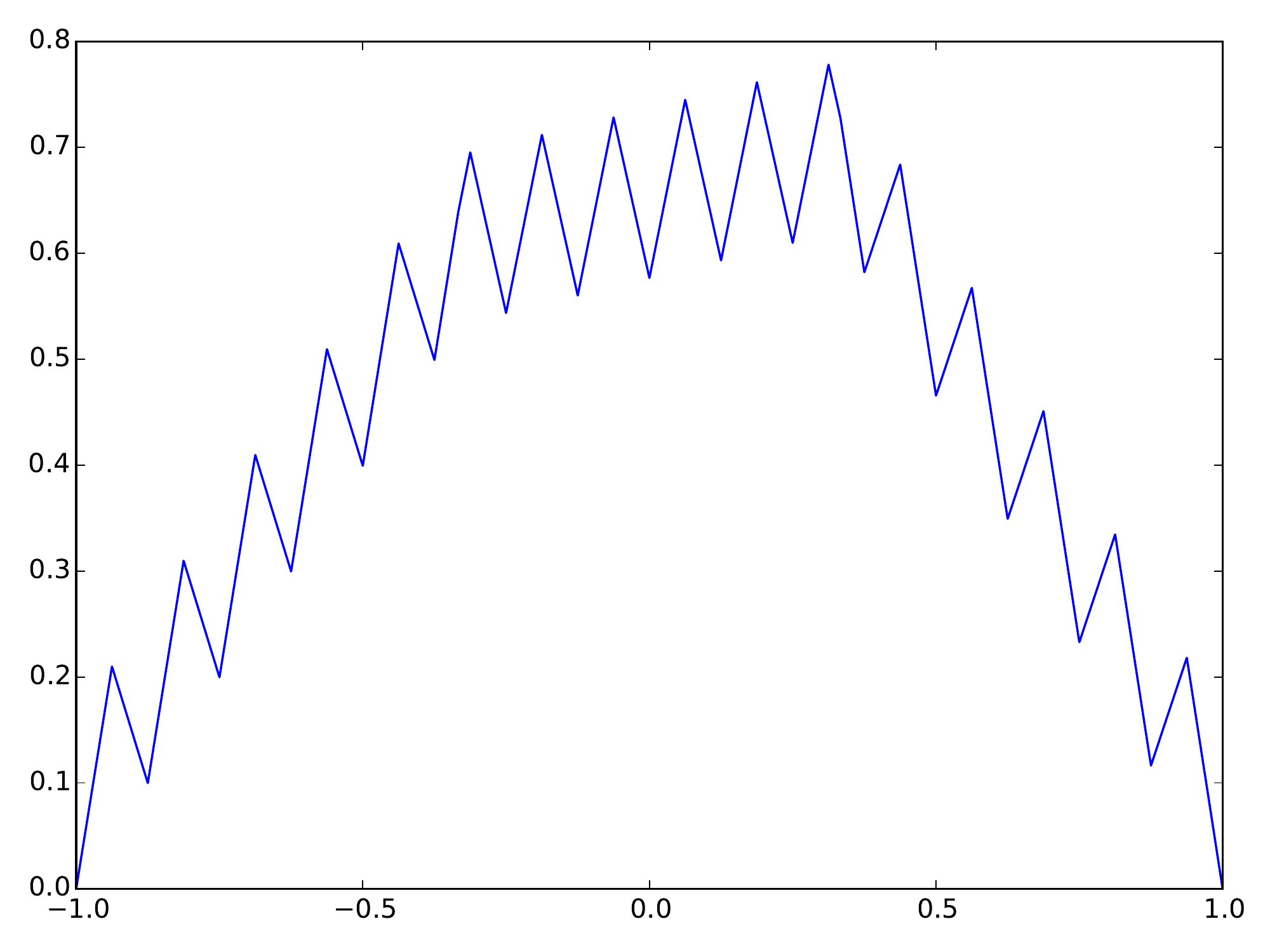}
\caption{Our main example model, constructed as a sum of a high-frequency
sawtooth wave and a low frequency spline (left) and the graph of
the output function of this network (right).  Our hand-constructed model
has nonzero weights only on the bold lines, but the models trained on its
output have no constraints on the weights.}
\label{fig:low_model}
\end{center}
\end{figure}

Note that because many nodes on the right are unused, even this
network is larger than it needs to be to perfectly fit the output function.
For our main working example, we chose a random $5\times (2+3)$ network
of this form.  We want to construct the smallest possible interesting example, 
to make numerical study easier, but networks much smaller than the one
constructed here tend to be degenerate. See Figure~\ref{fig:low_model} for 
the network structure and example function. We refer to this output function 
as $f:\R\to\R$ and the perfect-fit network weights as $W_f$.

Note, though, that when we train a $5 \times 5$ model to the output of
$W_f$, we do not enforce the two sided structure --- the network
is fully connected on each layer.  The $2+3$ format is just the hand crafted
way to produce the sum of high and low frequency components.

\section{Symmetrization}\label{sec:symmetrization}

If two weights $W_1$ and $W_2$ produce the same output function,
we say they are symmetric. The space of weights of a $K\times N$
network has a large group of symmetries.  In particular, it contains the
group $\R_+^{KN}\rtimes S_{N}^K$, where $S_N$ is the symmetric group on $N$
elements and $\R_+$ is the multiplicative group of positive reals.  To see
this, note that we can rearrange the indices on the hidden nodes at will,
producing the group $S_{N}^K$, and we can scale up the weights and bias at
a particular node, as long as we perform the reciprocal scaling on the 
associated weights in the next layer up, producing $\R_+^{KN}$, and
the action of $S_N^K$ on $\R_+^{KN}$ in the semidirect product
is the obvious permutation action.

The presence in the symmetry group of $\R_+^{KN}$ means that at any point $W$
in weight space, there is a $KN$-dimensional submanifold passing through $W$
along which the output function is completely unchanged, and the presence
of the $S_{N}^K$ factor means that there are a large number ($(N!)^K$) of
these submanifolds in weight space, each with identical outputs.  There is a 
question of whether removing all of these symmetries from the weight space 
will make training easier.  See~\cite{sym}, for example, for more background.

Because we are interested in exploring the actual weight space of
our network structure and not just the output functions, we introduce a symmetrization
procedure, as follows.  First, find the element of $\R_+^{KN}$ which minimizes
the $L_2$ norm of the entire weight vector (the ``minimal energy weights''), 
then permute the nodes in each layer so the columns of the weight matrices are in 
increasing $L_2$ norm order.

In our experiments, this procedure, applied after training, tends to 
make paths in weight space slightly smoother, but it does not appear 
to fundamentally affect any of the observations which follow. Our main 
purpose in symmetrizing is that our example network $W_f$ weights were
constructed by hand, and as such they tend to be rather different from
the initialization weights or the weights to which the network trains;
in particular, $W_f$ contains unusually large values. We want to be fair 
and explore the neighborhood of the ``generic representative'' of all 
networks symmetric to our example, in weight space, and we choose the 
minimum energy weights as our representative. Hereafter, $W_f$ refers 
to these minimum energy weights, and we find minimum energy 
representatives for all fitted networks.

We emphasize that this symmetrization does \emph{not} affect the training 
of the network: we do no quotienting of the weight space during the training 
process.
It is only afterwards, when we want to explore the weight space, that we choose
nice representatives of the fitted networks.

\section{Experimental results}
\label{sec:results}

\subsection{Initial example fit}

\begin{figure}[ht]
\begin{center}
\includegraphics[scale=0.4]{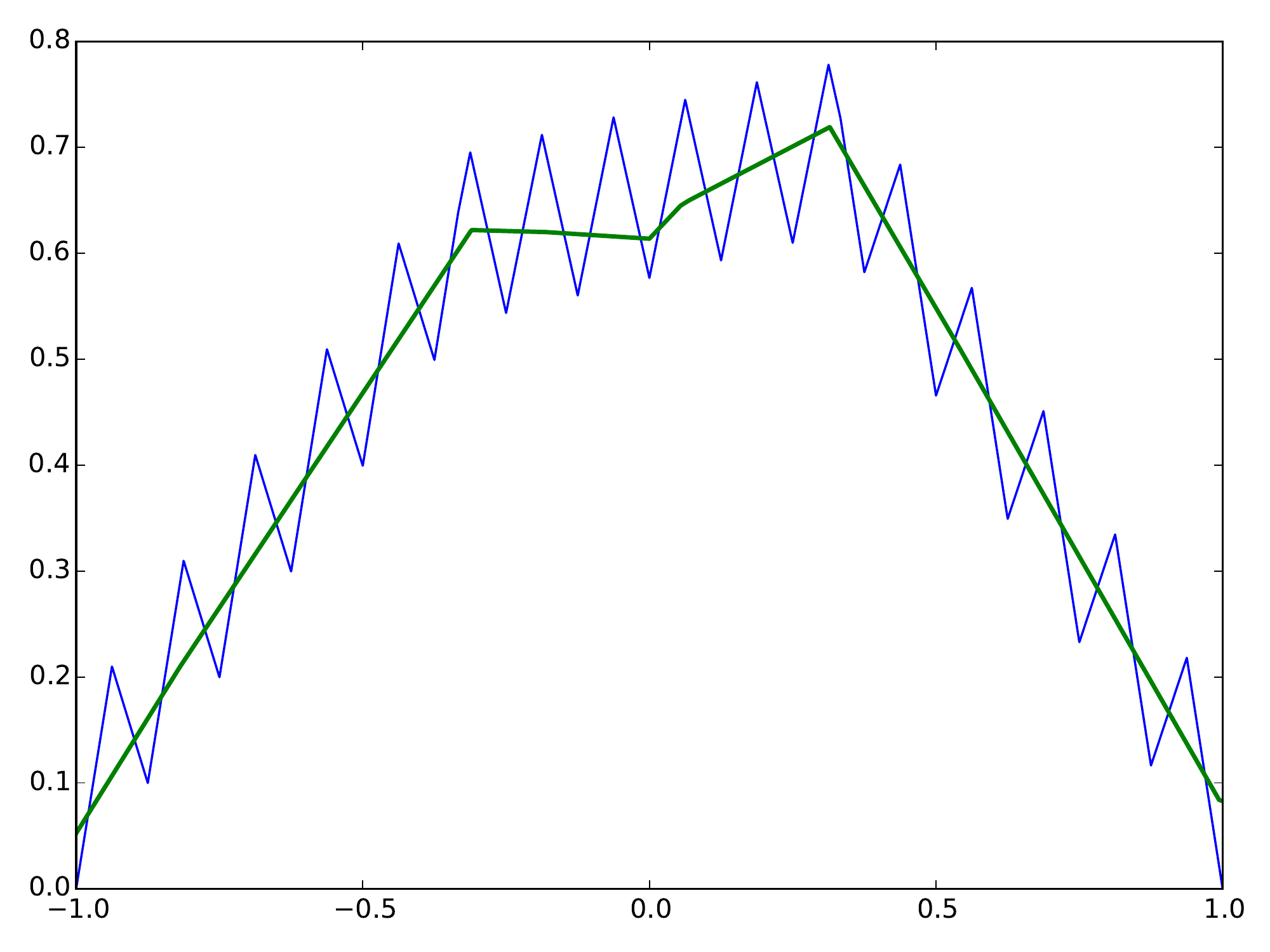}
\caption{Fitting a network of the same size to the output $f$ of $W_f$ (blue)
always produces a network with output similar to the green plot.}
\label{fig:fitted}
\end{center}
\end{figure}

First, we give an example plot showing our main theme.
If one fits a $5\times 5$ network to the function $f$ and trains
until the weights appear to stabilize, the fitted
network will have output which looks similar to Figure~\ref{fig:fitted}.
It doesn't much matter which optimizer one uses or how long one trains.

\subsection{Fitting method}

All of the models discussed in this paper were trained in the following
way, unless otherwise mentioned.
Because we know the function we want to fit, exactly, we are in the 
unusual position of having as much data as we'd like.  Our training data
consisted of the set of all knots in the linear spline of $f$, plus $9$
points on each of the linear segments between knots; this produces training
sets of size $n = 32\times 10 = 320$ points for our principal example. 
Experimentally, once we provide 2 to 3 points per segment, changing the amount 
of data has no effect on the results. Indeed, after enough data is provided,
the complexity of a trained network is a function of the associated energy landscape.
On each training step, 
we provided this entire dataset to the model, and we used either gradient 
descent with momentum or adadelta for 50,000 steps.

In practice, it takes only a few hundred steps until the model appears to
stabilize with either optimizer.
At this point, running gradient descent for tens of thousands more
steps only serves to emphasize that we have really found a local minimum.
Adadelta typically does a little better when given such a large number of
steps because it is willing to walk around in a slightly more random fashion.
Thus, it is more likely to stumble closer to the global minimum.

When doing our study of local minima, we chose to use the minima output
from gradient descent.  We wanted to make sure that at the end of training, we were
truly at a numerical local minimum, rather than in the middle 
of a ``clever'' sidestep.  We found that when starting at the output of
an adadelta fit, minimal energy paths in weight space between fitted
models tended to initially make a (very tiny) sharp descent in energy,
so we felt uncomfortable calling these fits local minima.
In general, we feel the gradient descent optimizer gives us a more
reasonable sampling of the space, and the adadelta optimizer represents
a comparison with a modern method that involves more randomness and is
trying quite hard to succeed in any way it can.

For our main dataset, we fitted 320,000 $5\times 5$ networks as above,
half with gradient descent and half with adadelta.

\subsection{Fitted plots}

\begin{figure}[ht]
\begin{center}
\includegraphics[scale=0.34]{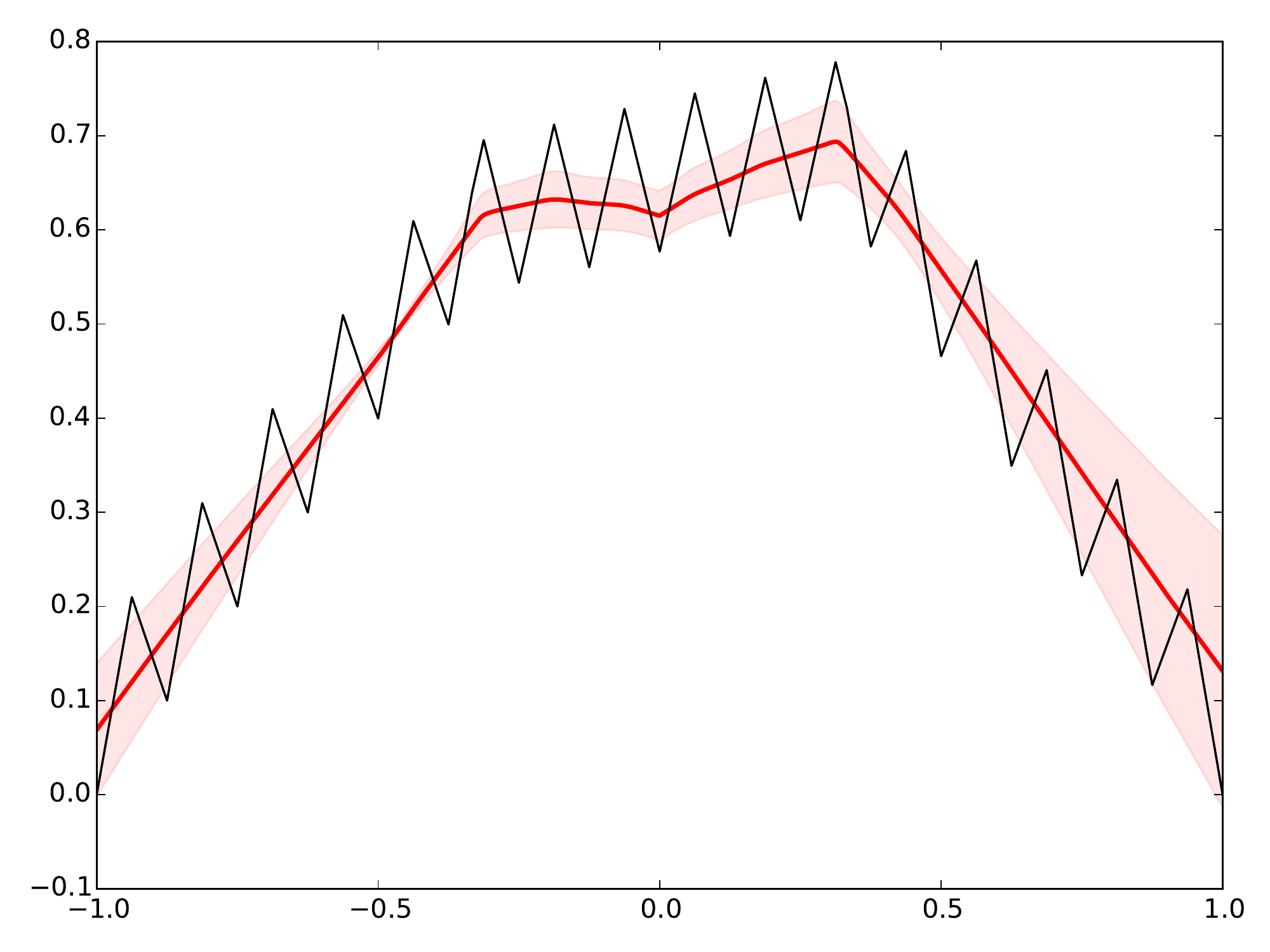}\includegraphics[scale=0.34]{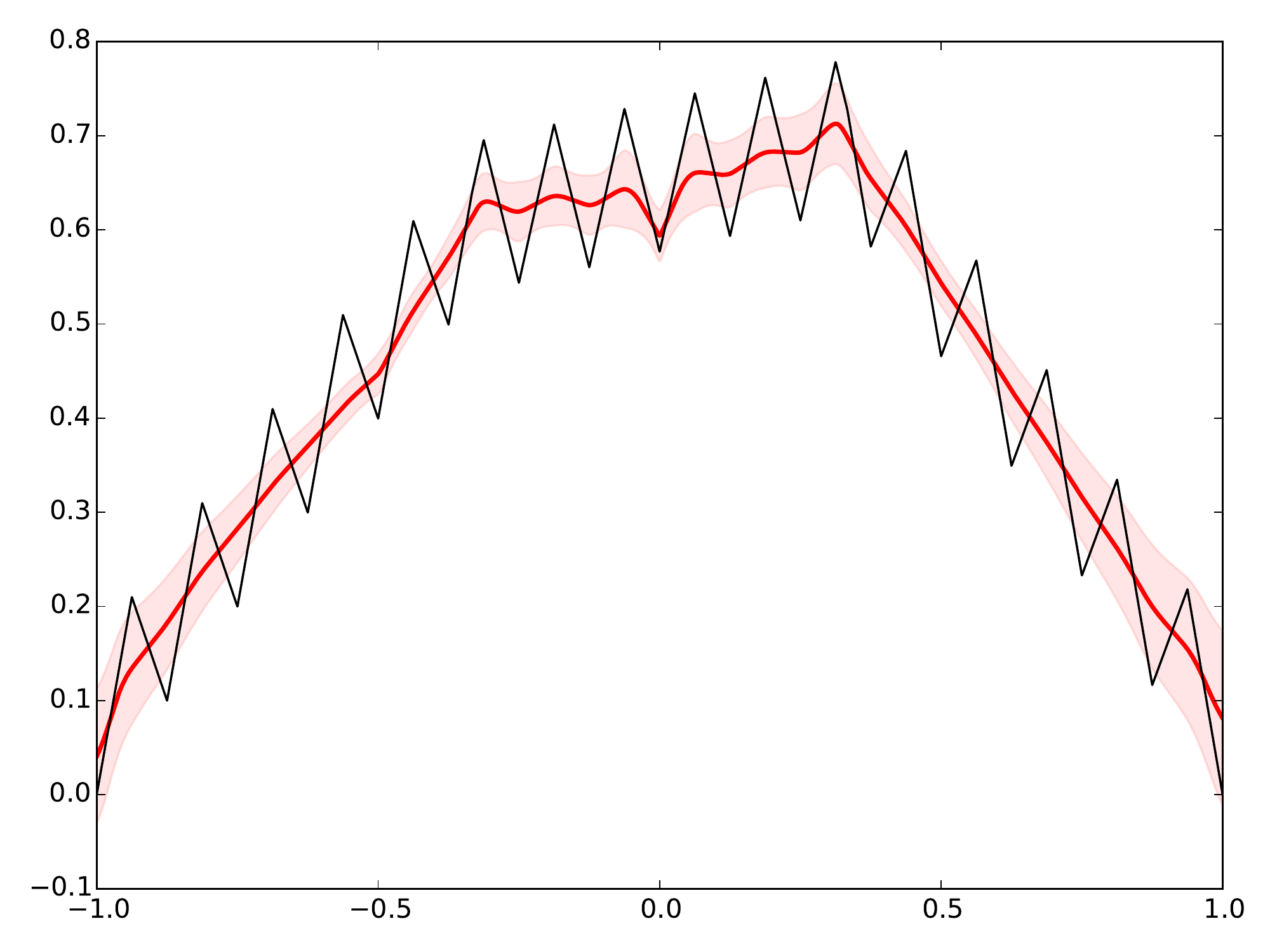}
\caption{The average fit and standard deviation tubes for gradient
descent and adadelta over 160,000 trials.}
\label{fig:average_fits}
\end{center}
\end{figure}

We claimed above that essentially any fit of a $5\times 5$ network
to the function $f$ would produce the same smoothed output without
the high-frequency sawtooth, and certainly wouldn't find the global
minimum.  Figure~\ref{fig:average_fits} shows the average output of the trained
networks and a tube around the average showing the standard deviation for both
optimizers.  Note that the average fit for adadelta has a slight wiggle
in the middle, but in general, the averages are quite flat, indicating
that neither method can find the high frequency component.

\begin{figure}[ht]
\begin{center}
\includegraphics[scale=0.34]{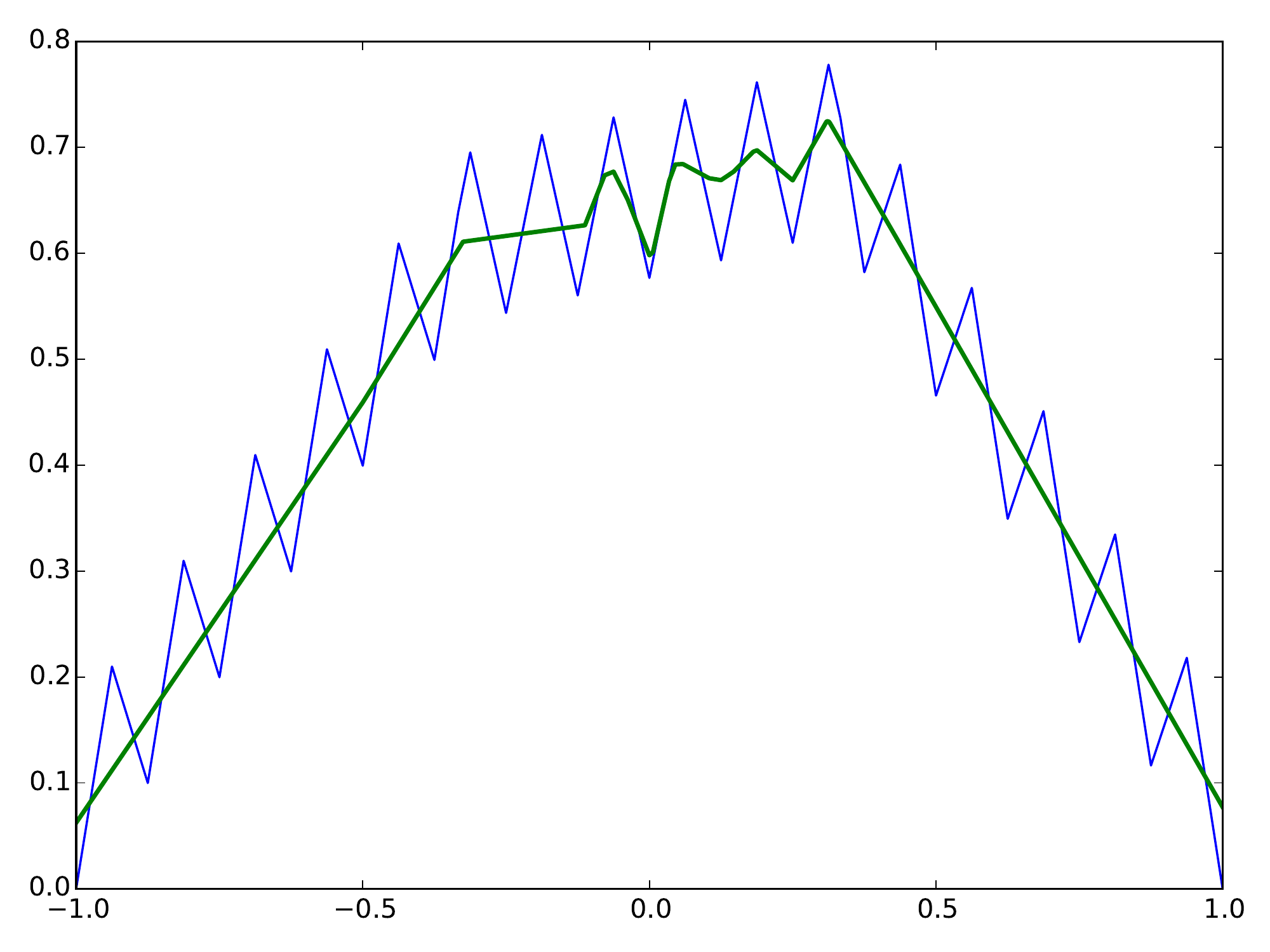}\includegraphics[scale=0.34]{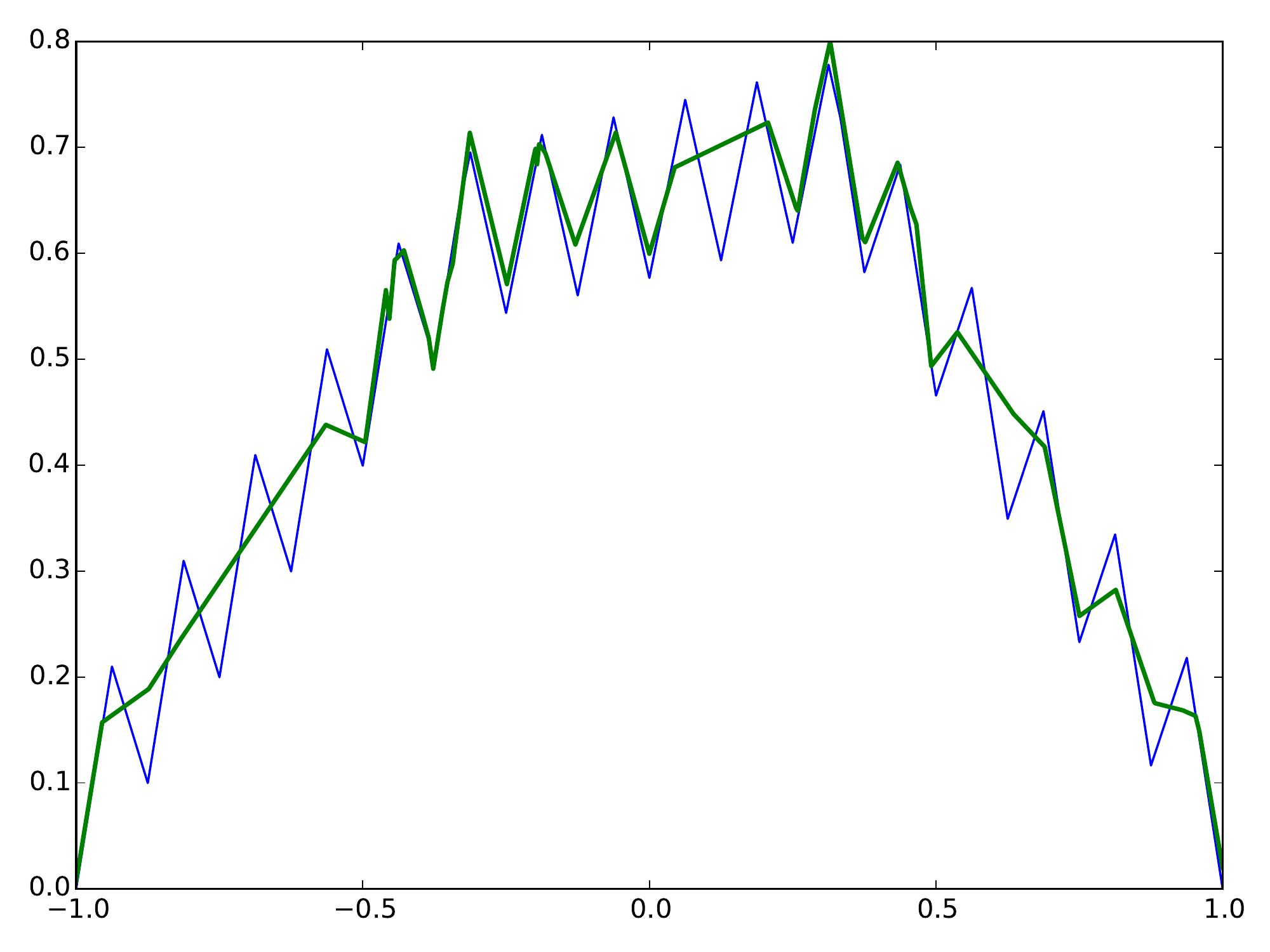}
\caption{The best fits for gradient descent and adadelta over 160,000 trials.}
\label{fig:best_fits}
\end{center}
\end{figure}

Figure~\ref{fig:best_fits} shows the absolute best fits over 160,000 trials
each.  Clearly, adadelta does better, but it can't quite find an optimum.

\subsection{Losses}

\begin{figure}[ht]
\begin{center}
\includegraphics[scale=0.4]{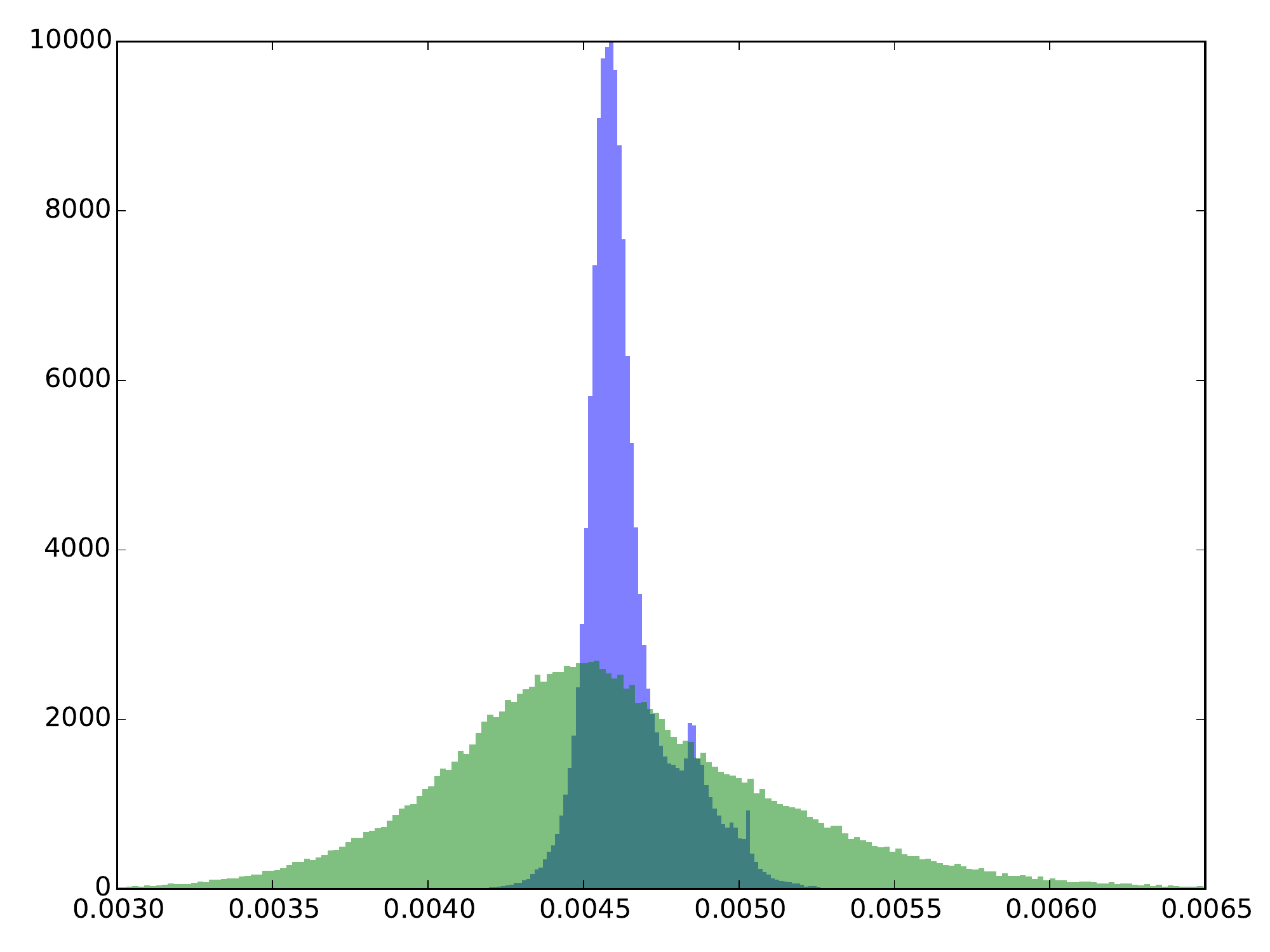}
\caption{Histograms of the mean squared errors of gradient descent (blue) and adadelta (green) fits.}
\label{fig:loss_hists}
\end{center}
\end{figure}

The mean squared error loss corresponding to the average (smooth, low-frequency)
fit is about $0.0045$.
Figure~\ref{fig:loss_hists} shows histograms of the losses over the 320,000
fits for both gradient descent and adadelta.  Note the scales on the horizontal axes
are quite small, so the qualitative difference between the distributions is exaggerated.
Adadelta does better, but it has a wider variance, and both distributions
have a large spike at the smooth fit.  We do not have a good explanation for the specific features of either distribution, but it is clear that the barriers
between the local and global minima are too high for either technique to climb
over, and the energies associated with the individual local minima found by
either technique are fairly similar.

\subsection{Distances between minima}

\begin{figure}[ht]
\begin{center}
\includegraphics[scale=0.4]{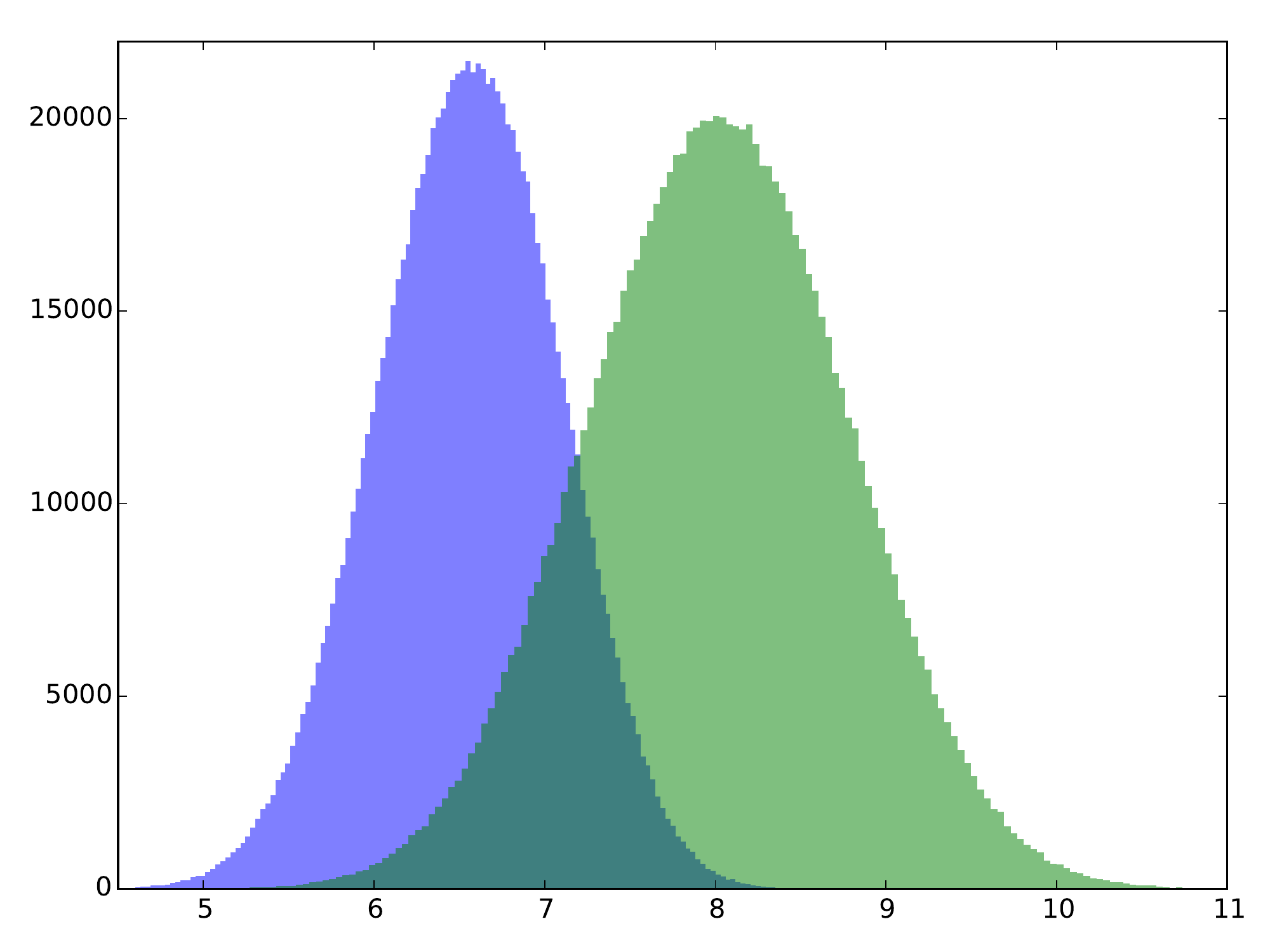}
\caption{Histograms of the $L_2$ distances between pairs of weights of fitted models, with gradient descent (blue) and adadelta (green).}
\label{fig:pair_dists}
\end{center}
\end{figure}

We might ask: do we ever train to the same point?  As discussed above, there
are entire submanifolds of weight space which produce exactly the same
output (so the set of local minima is not discrete), but we might hope
that the symmetrization procedure from Section~\ref{sec:symmetrization}
would remove this issue, and after symmetrization we might find duplicate minima.
In fact, we do not.  Figure~\ref{fig:pair_dists} shows histograms of
the $L_2$ distances between $1,000,000$ randomly chosen pairs of
trained weights.  The minimum distances are both about $4$.

\begin{figure}[ht]
\begin{center}
\includegraphics[scale=0.4]{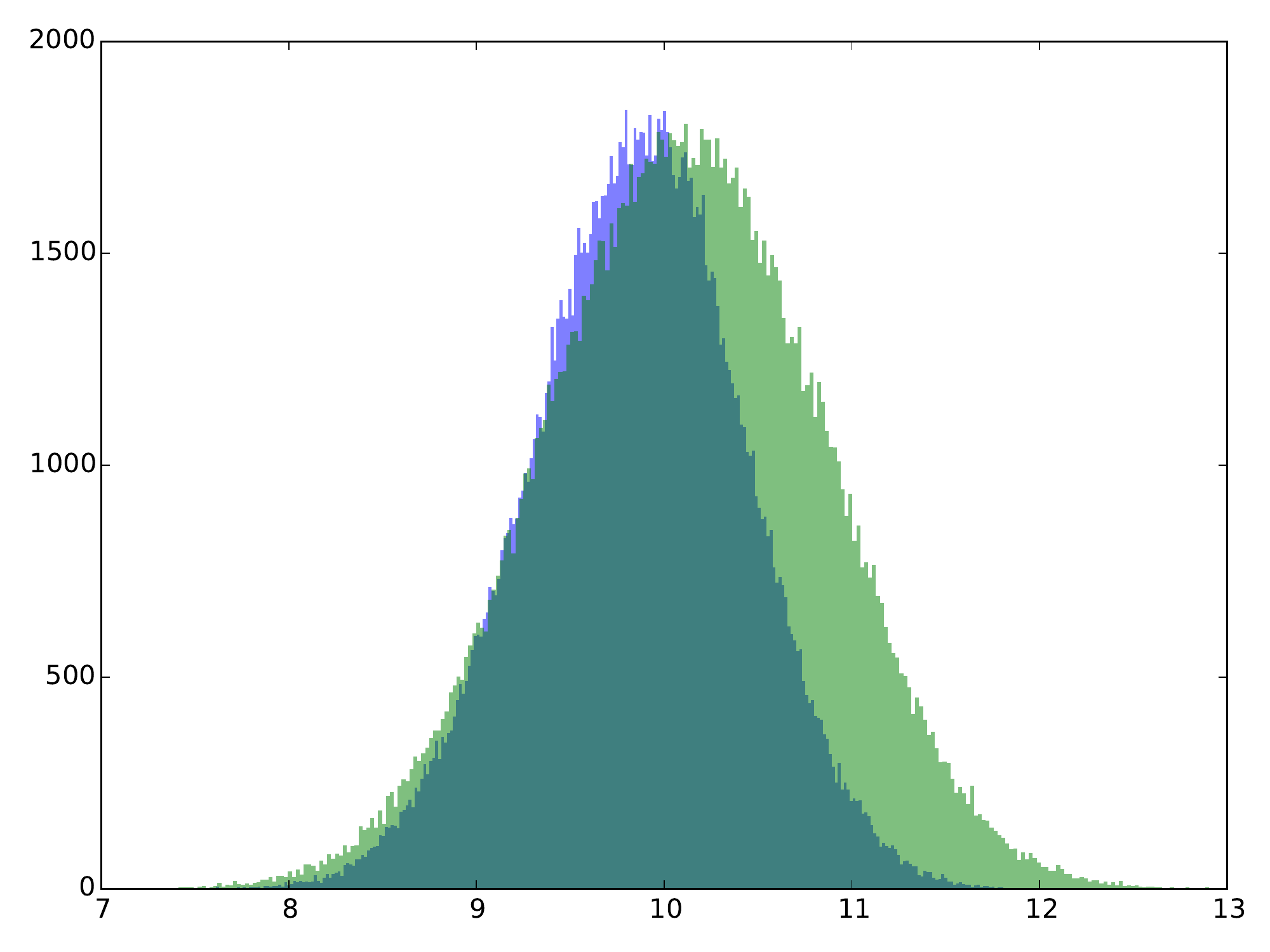}
\caption{Histograms of the $L_2$ distances between the weights of fitted models and a global minimum,
with gradient descent (blue) and adadelta (green).}
\label{fig:min_dists}
\end{center}
\end{figure}

We can also compute the distances between a global minimum
(the handcrafted weights, after symmetrization) and all the local minima.
A histogram of these distances is shown in Figure~\ref{fig:min_dists}.

It may seem somewhat paradoxical that all the pairs of local
minima are within about distance 6-8 of each other, but they are
also all about distance 10 from a global minimum.  The only way for this to occur is
if they are all to one side of the minimum, which seems counterintuitive.
In fact, this is exactly the arrangement.  Suppose we compute the direction vectors
from the global minimum to each local minimum and take dot products of pairs of these vectors.
If the local minima were distributed equally around the global minima, we would expect
to find a range of dot products (including some negative ones).  In fact, the minimum dot
product is about $0.6$, indicating that the global minimum is off in a corner of the space.
Note that this is not particularly surprising in high-dimensional space: a random
Gaussian collection of points in $d$-dimensional space will have the same property,
as long as there are fewer than exponentially (in $d$) many points.

\subsection{NEB paths}

\begin{figure}[ht]
\begin{center}
\includegraphics[scale=0.34]{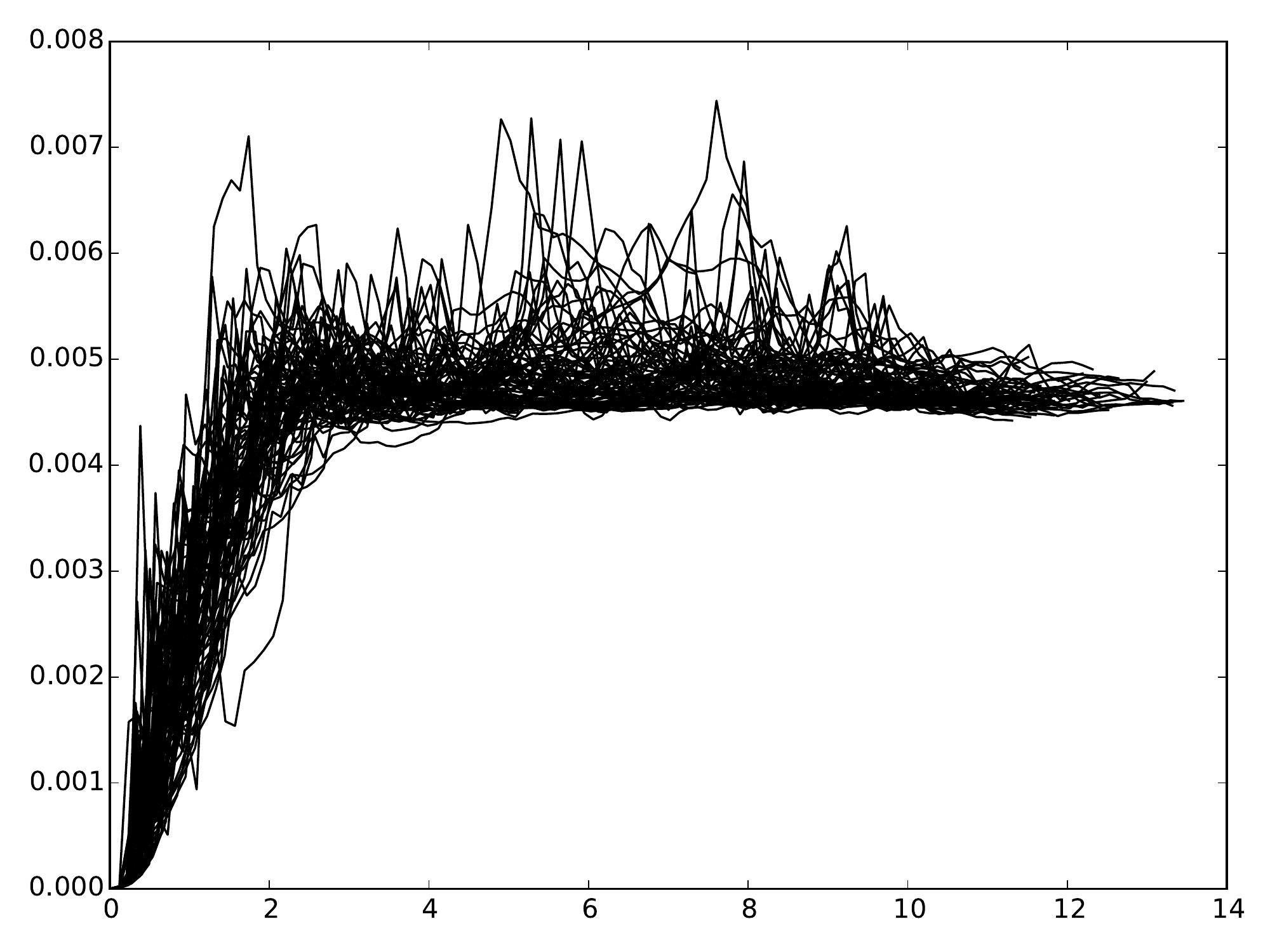}\includegraphics[scale=0.34]{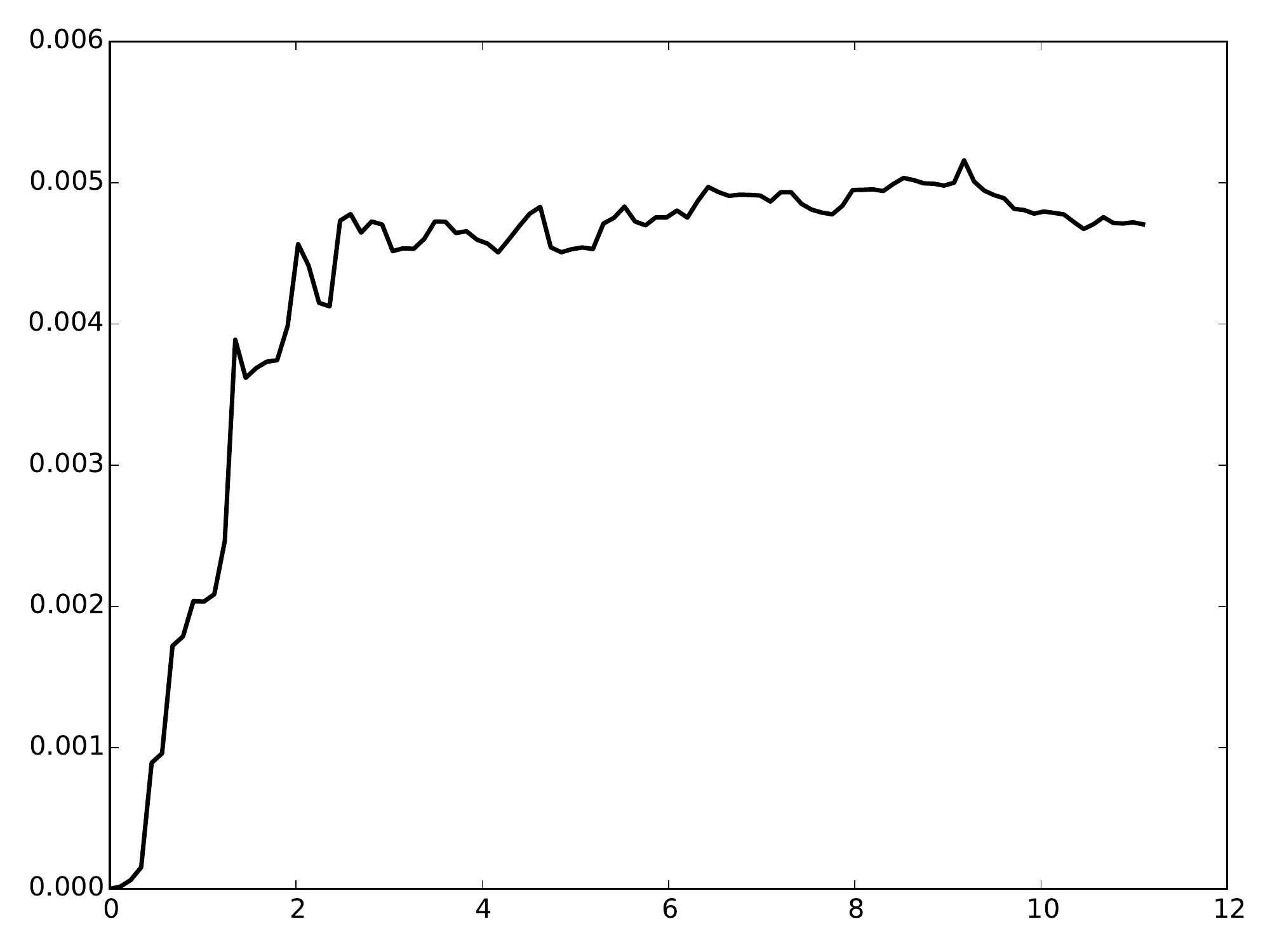}
\caption{One hundred NEB paths from the global minimum to local minima (left),
showing the very flat average terrain, and a single example NEB path (right).
Note that in 136 dimensional weight space, the volume of the unit ball centered
at the global minimum is much smaller than the volume of the shell of radius
$9.5$ to $10.5$ which contains most of the local minima. The apparently gentle
slope to the global minimum is an artifact of the one-dimensional NEB path and
the chance of initializing the network to the domain of attraction of the global
minimum is essentially $0$.}
\label{fig:NEB_paths}
\end{center}
\end{figure}

Our goal is to gain some sense of the topography of the energy
landscape.  This is difficult, as the weight space has 136 dimensions.
However, we can try to understand the landscape as we travel
from a global minimum to a local minimum along a minimum energy path.
This gives an idea of the terrain that the network must traverse
under training if it is to find the global minimum.
We used the nudged elastic band (NEB) method (see~\cite{NEB}) to find these minimal energy paths.
These NEB paths demonstrate that the energy landscape is flat and slightly
bumpy until a dramatic dropoff near the global minimum.
See Figure~\ref{fig:NEB_paths}.  Such a landscape is extremely
difficult for any gradient descent-based method to traverse.

\begin{figure}[ht]
\begin{center}
\includegraphics[scale=0.34]{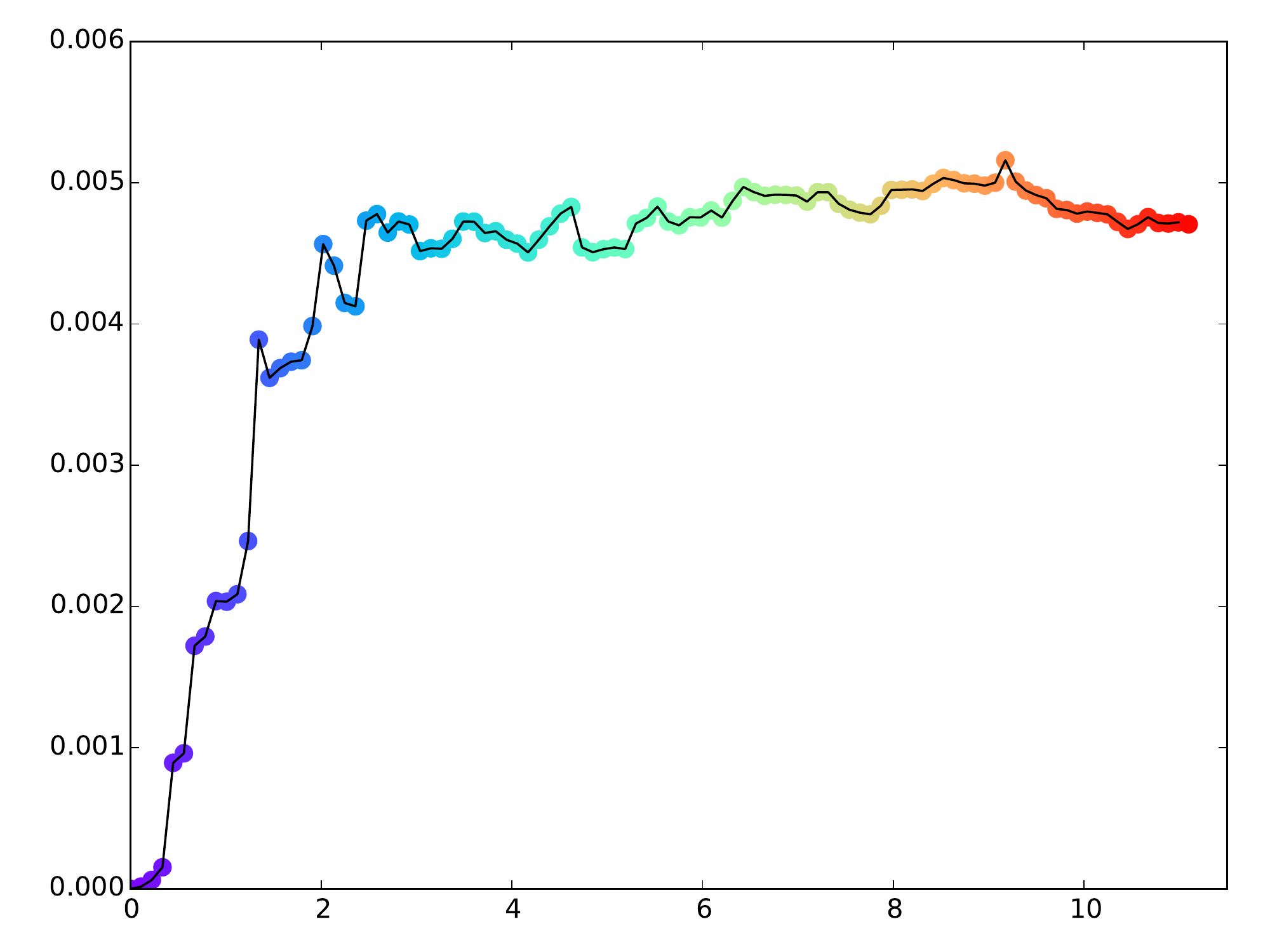}\includegraphics[scale=0.34]{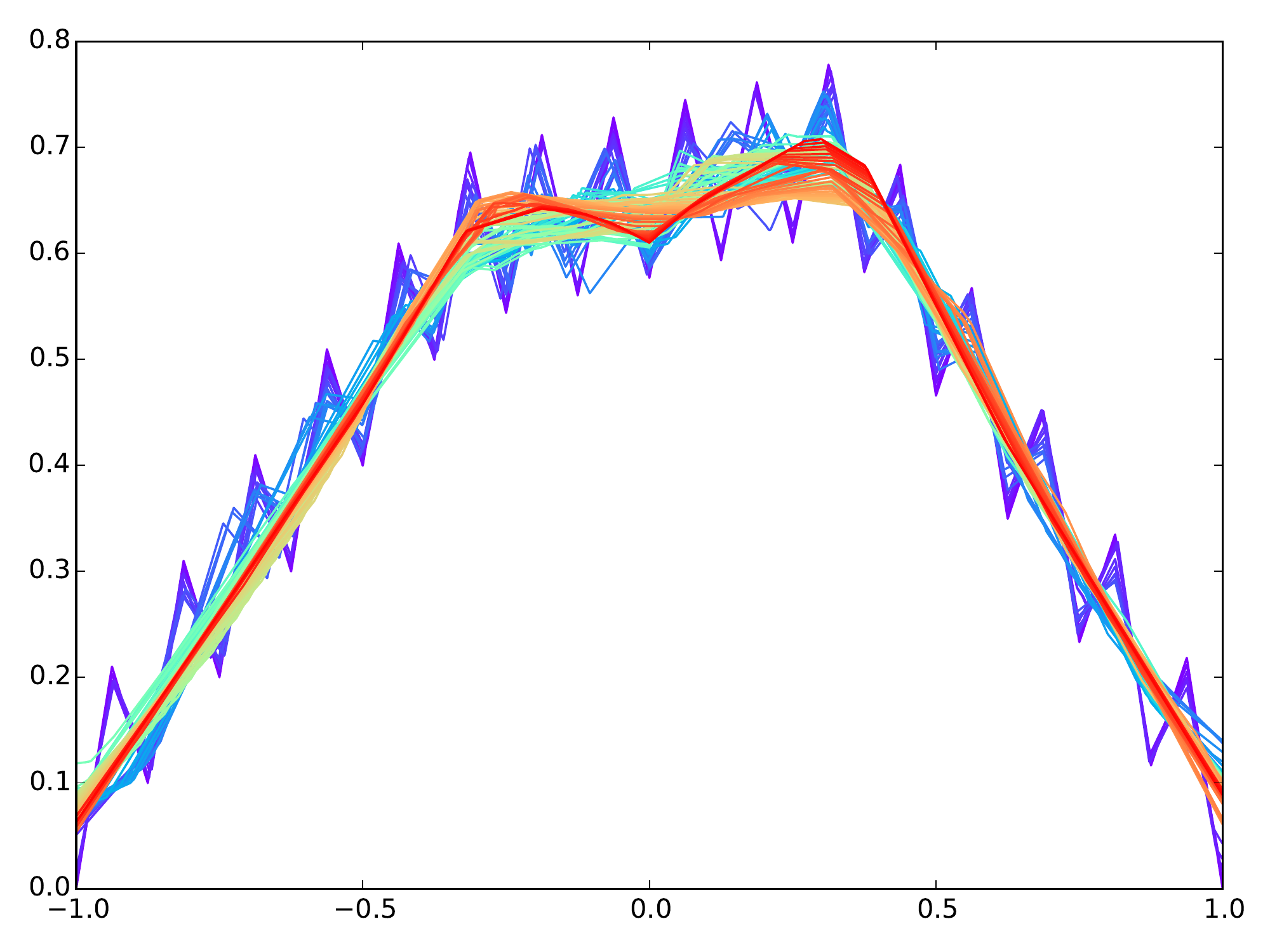}
\caption{A single NEB path and the corresponding network output along the path.}
\label{fig:NEB_path_plot}
\end{center}
\end{figure}

Although it is not particularly surprising, it is interesting to observe what
happens to the output function as we travel along a NEB path.  During the large, flat
region of the path, the output changes very little as the function rearranges itself
while preparing to create many spikes, which it does at the very end as it
plunges towards the global minimum.  See Figure~\ref{fig:NEB_path_plot}.

\begin{figure}[ht]
\begin{center}
\includegraphics[scale=0.34]{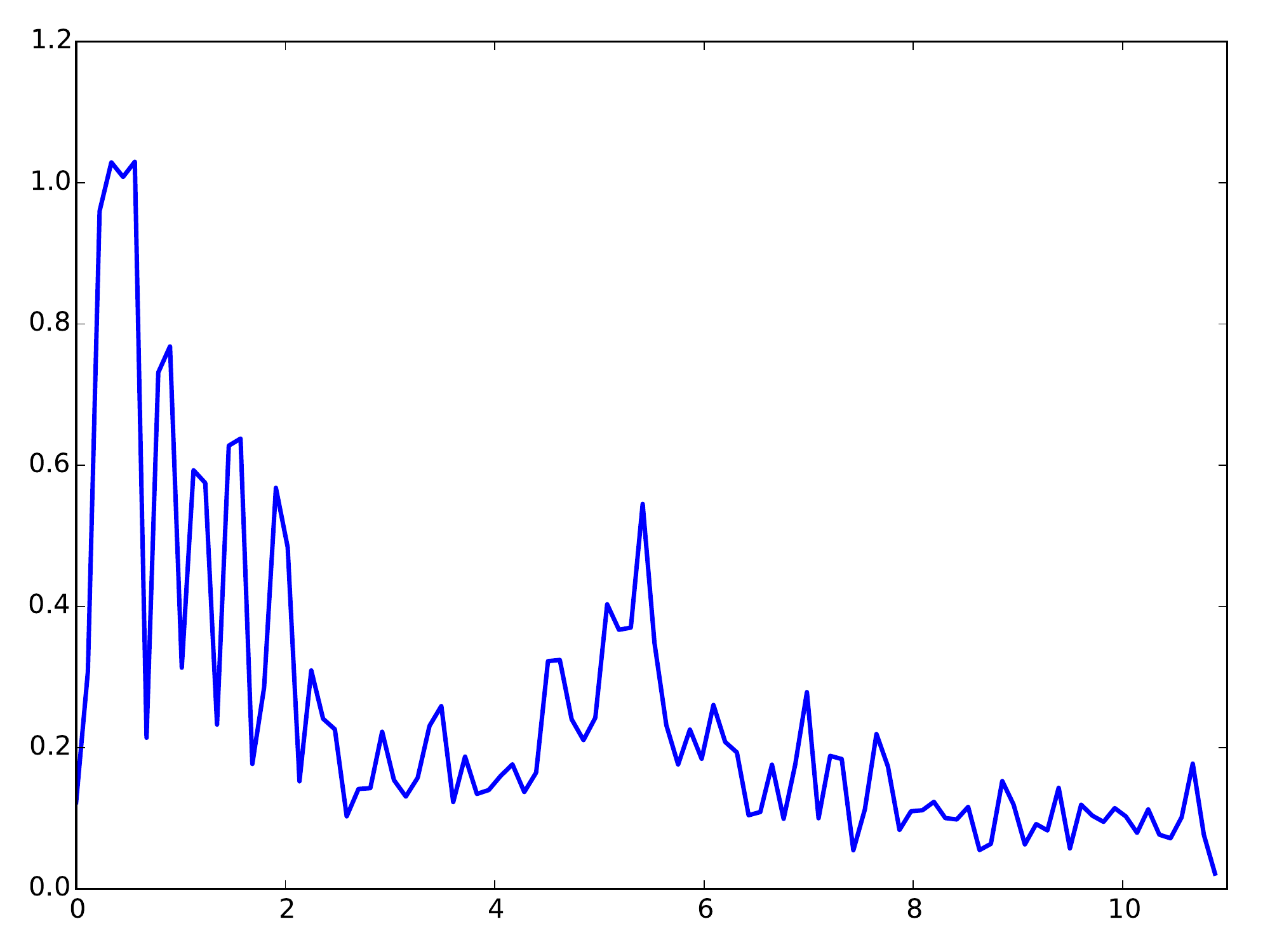}
\caption{The angles (in radians) between successive difference vectors as we travel along the NEB path.
The horizontal axis is distance along the path.
Near the global minimum, the path twists and turns a lot.}
\label{fig:NEB_path_angle}
\end{center}
\end{figure}

Finally, we can get an idea of how complicated the energy function is near
the global minimum by plotting the angles between successive difference vectors
along a NEB path.  Figure~\ref{fig:NEB_path_angle} gives an example
of this.

As we expect, the path is required to do some twisting
if it wants to approach the global minimum in an energy-minimizing way.
We remark that we originally linearly interpolated between each pair of points
along the NEB paths for plotting to increase their resolution, but we found
that the interpolated points were useless: the energy function has such sharp curved
valleys that the interpolated points have much larger loss.

\subsection{Eigenvalues}

\begin{figure}[ht]
\begin{center}
\includegraphics[scale=0.7]{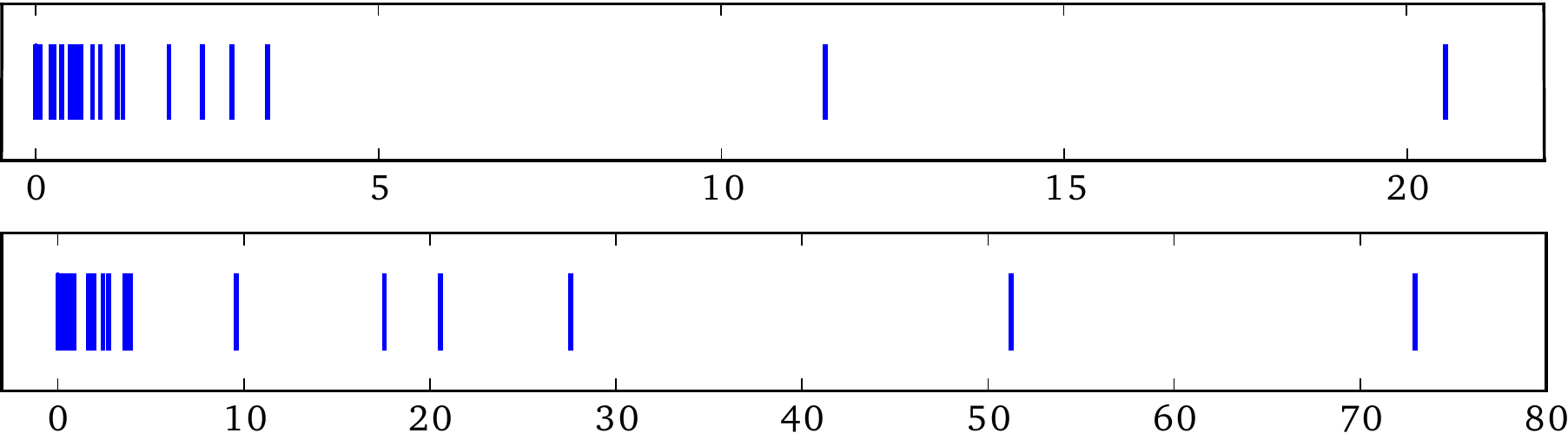}
\caption{The eigenvalues of the Hessian of the loss function
at local (top) and global (bottom) minima.  Note there are 136 eigenvalues drawn;
most of them are very close to $0$.}
\label{fig:eigenvalues}
\end{center}
\end{figure}

To gain a further insight into the energy landscape near the minima,
we can compute the eigenvalues of the Hessian of the loss.
The use of the relu activation function means that the loss is not
technically even $C^1$, but we expected that the single isolated issue at $0$
wouldn't actually affect the computation of the Hessian.
In fact, it does, and the numerical
Hessian is neither symmetric nor stable.  To avoid this issue, during the
computation of the Hessian, we changed the activation function
to $(1/5000)\log(1+\exp(5000x))$.  This function, while smooth,
is close enough to relu that the output of the network is not
measurably changed.
Figure~\ref{fig:eigenvalues} shows that at both kinds of minima, all eigenvalues
are positive, and most of them are very small.

\subsection{Basin of the global minimum}

\begin{figure}[ht]
\begin{center}
\includegraphics[scale=0.34]{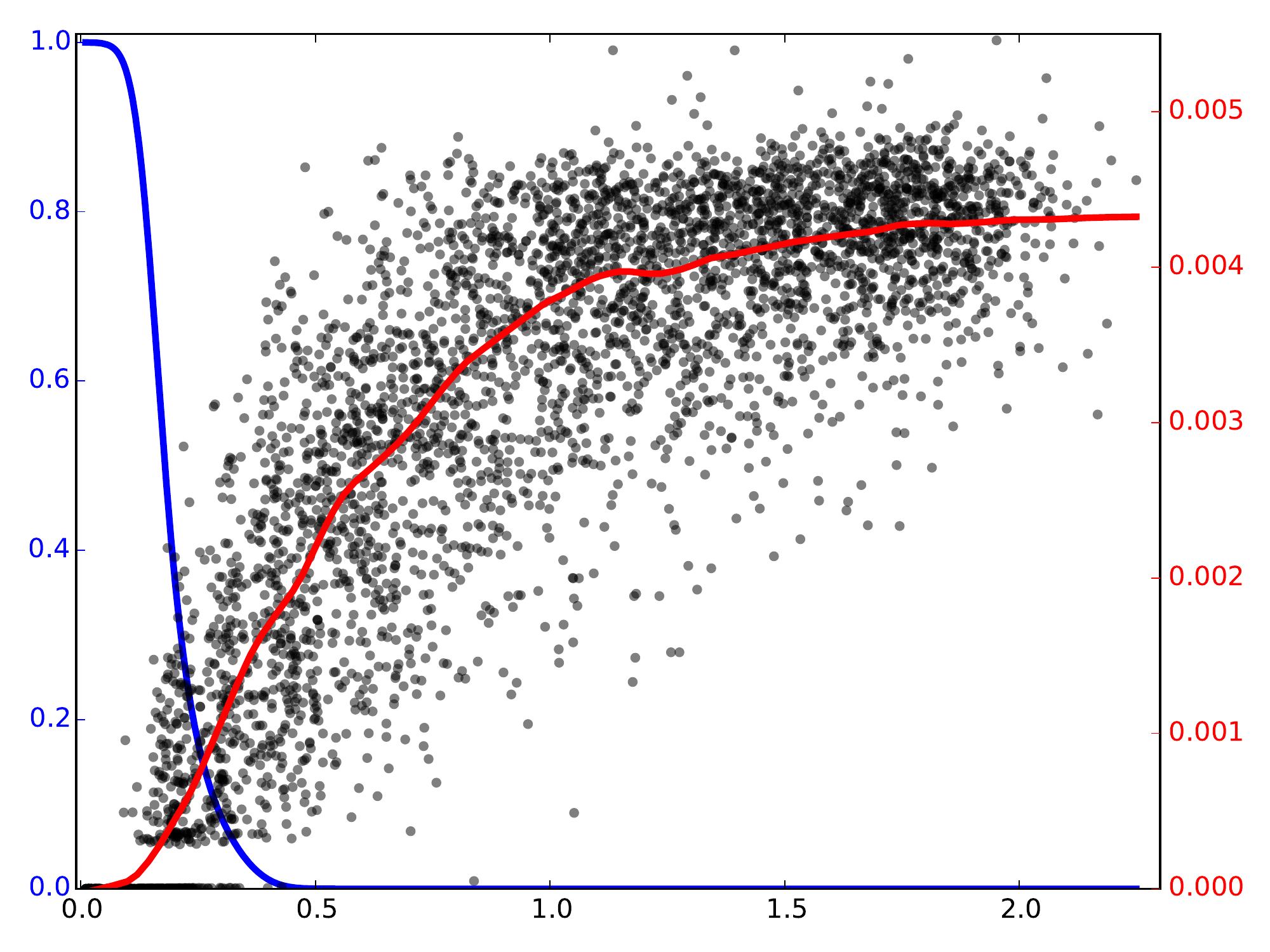}\includegraphics[scale=0.34]{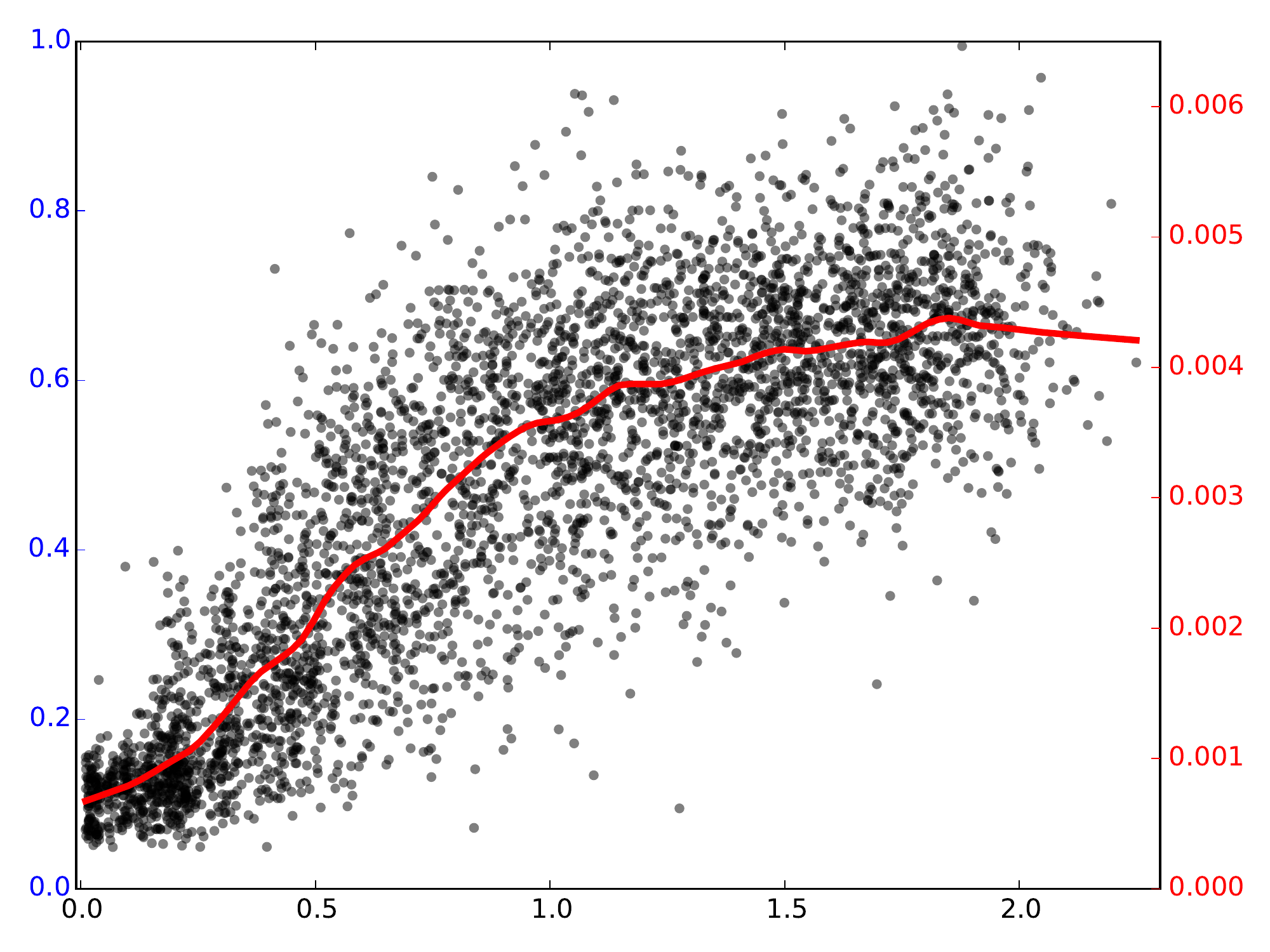}
\caption{For both gradient descent and adadelta, a scatter plot of the final
loss plotted as a function of the $L_2$ norm of the added noise
(initial distance), and a lowess approximation (red).
Also plotted is a logistic fit to the probability (blue) of returning to a global
minimum as a function of the added noise amount.  Note the two different vertical scales, and
note that the probability curve for adadelta is not visible because it is identically zero.}
\label{fig:jitter}
\end{center}
\end{figure}

The obvious initial rise of the NEB paths gives us an indication of the size of
the basin of attraction of the global minimum, but we can measure it another way:
if we jitter the weights of the global minimum by a small amount and then retrain,
we can find how much noise the model can take before it cannot find the global minimum again.
Figure~\ref{fig:jitter} shows the results of doing this experiment 4096 times
with gradient descent and adadelta.

For each experiment, we added a small amount of noise (with $L_2$ norm chosen approximately
uniformly in $[0,2]$) and trained for 20,000 batches, just as with our
main experiment above.  We then compute the loss of the final model and plot it against
the amount of noise added.  It is strange that adadelta seems to never find a perfect fit.
It is possible that it is doing small random walks, so stopping at a particular time is unlikely
to yield an exact minimum.  This agrees with our experience above.

\subsection{Another example}

\begin{figure}[ht]
\begin{center}
\includegraphics[scale=0.34]{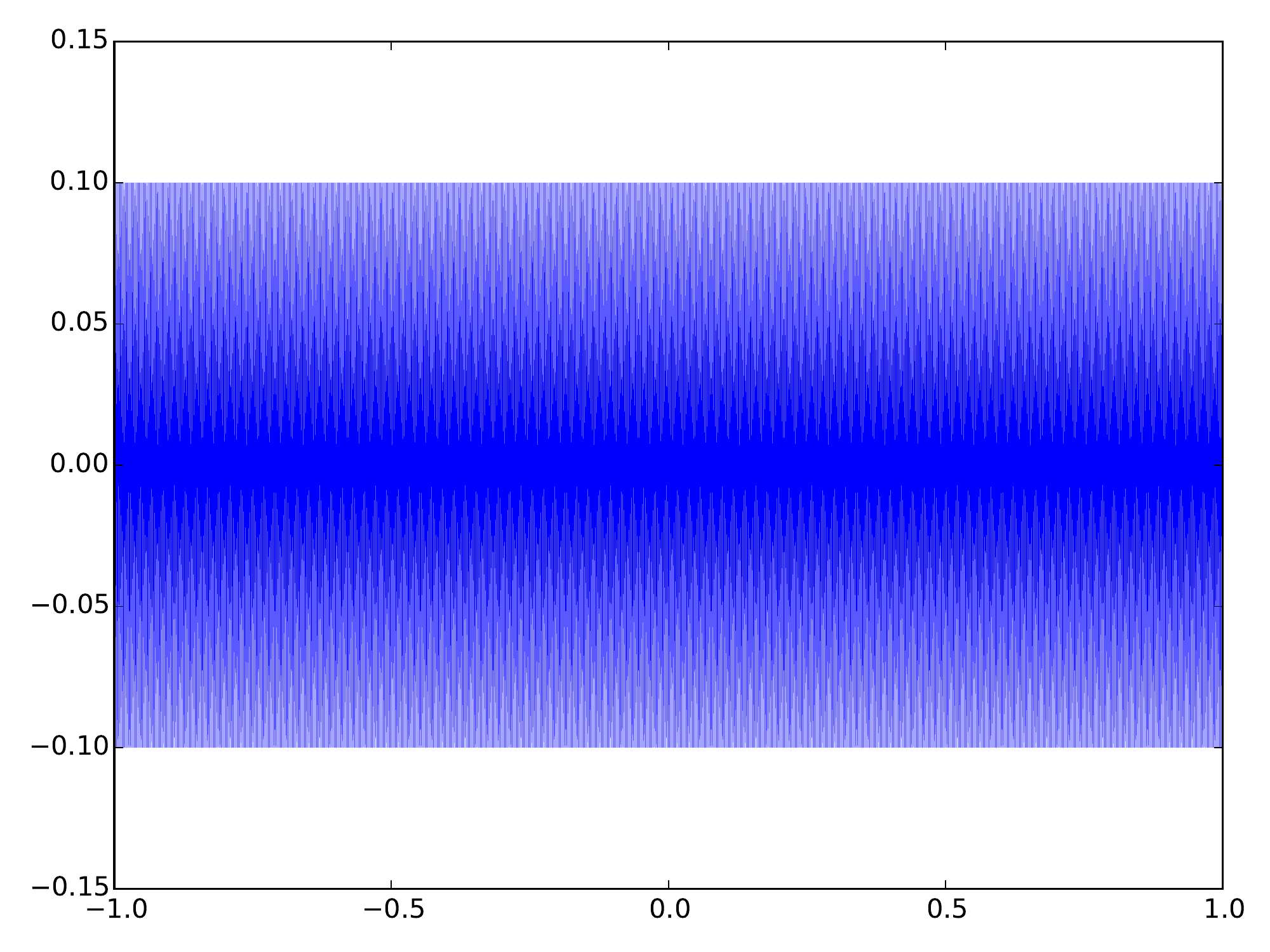}\includegraphics[scale=0.34]{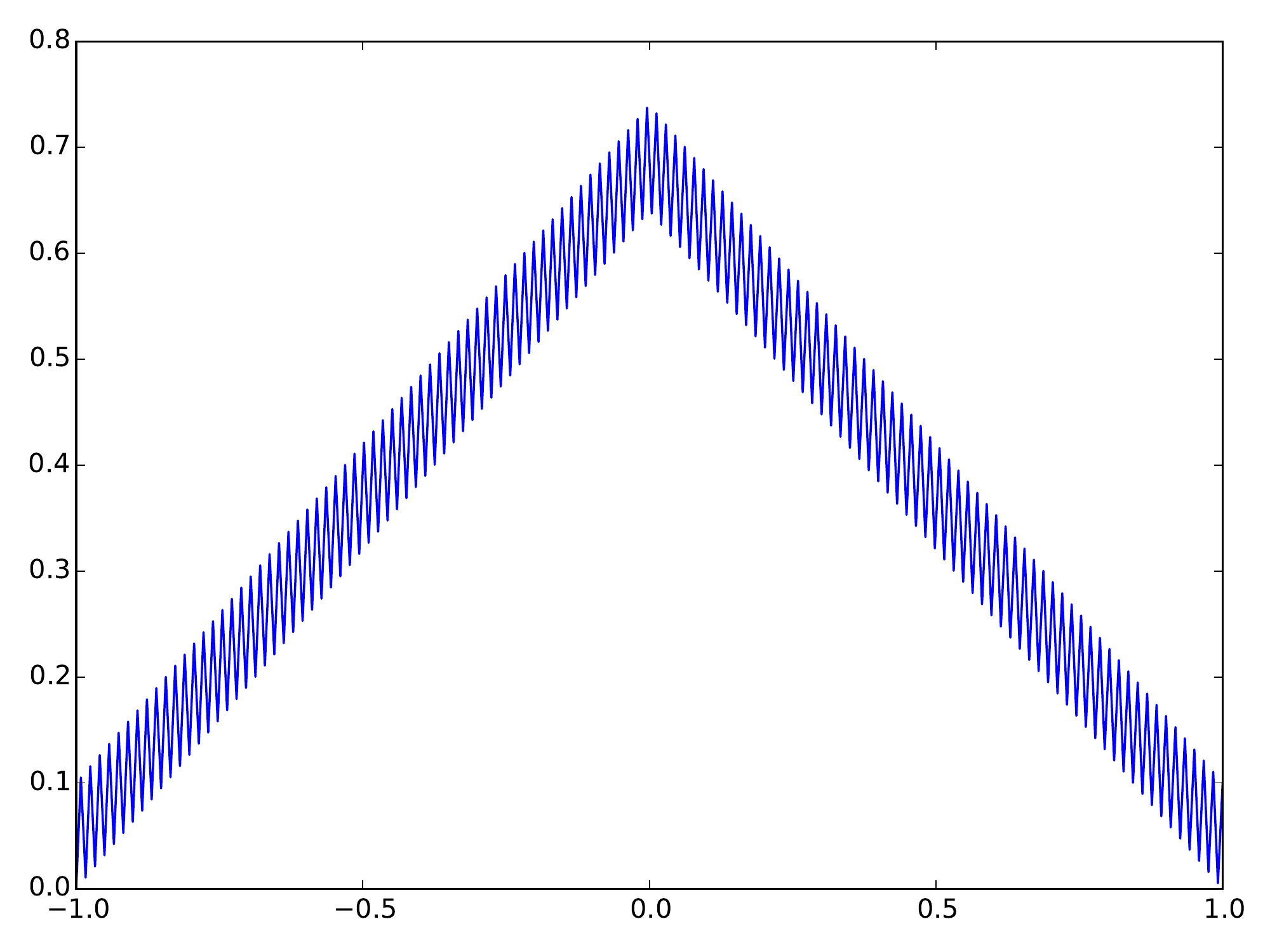}
\caption{A $5 \times (5+0)$ sawtooth (left); there are 3125 peaks and valleys,
and a $5\times (3+2)$ model (right).}
\label{fig:complex}
\end{center}
\end{figure}

Our main claim is that there are many functions which neural
networks of a given size could reproduce perfectly but will never learn
under training.  Our main example is actually a very gentle
one which does not even come close to the full complexity that
a $5\times 5$ network can create.  We now give some examples
demonstrating this fact.  See Figure~\ref{fig:complex}.

\begin{figure}[ht]
\begin{center}
\includegraphics[scale=0.34]{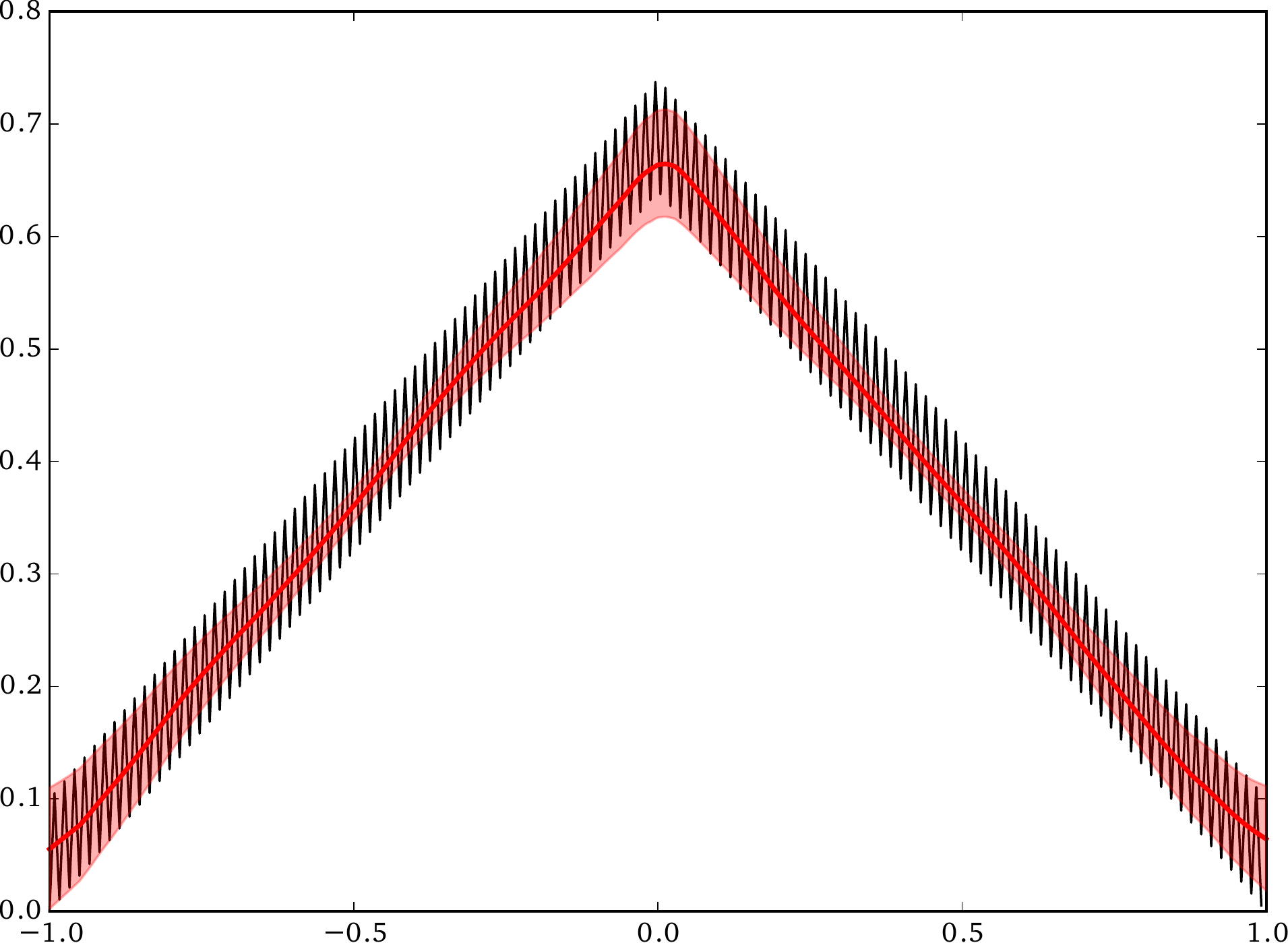}
\includegraphics[scale=0.34]{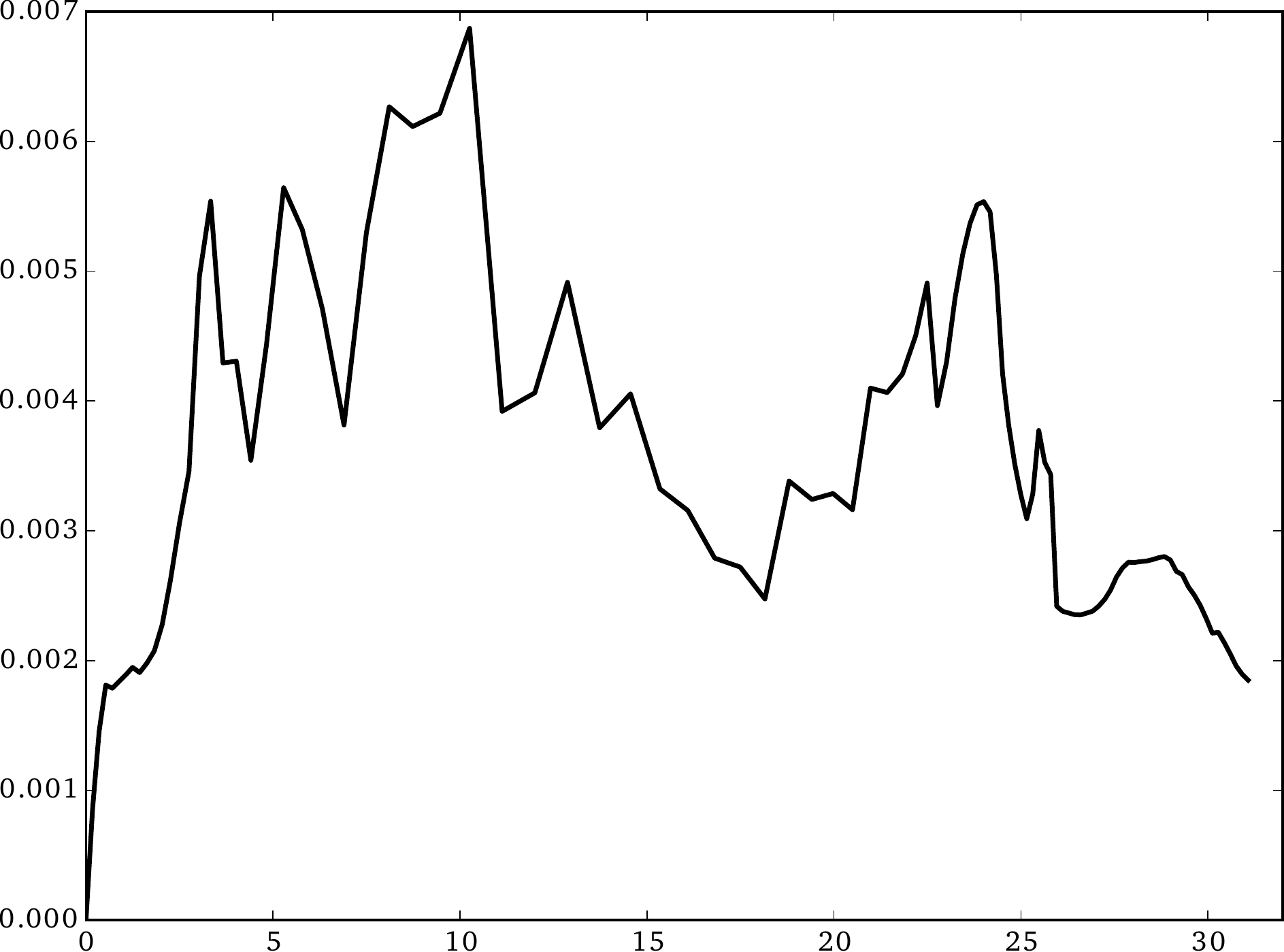}
\caption{The average fit and standard deviation tube for 256 adadelta $5\times 5$ fits
to the $5\times (3+2)$ model in Figure~\ref{fig:complex}, and a NEB minimal
energy path between the global minimum and one of these fits.  Note the deviation
in average fit plot is not due to some models fitting the function and some not: they all
produce wiggle-free fits, but the exact location of the fitted function varies.}
\label{fig:average_fits_complex}
\end{center}
\end{figure}

Obviously we expect to have trouble fitting such models with $5 \times 5$
networks.  Figure~\ref{fig:average_fits_complex} shows an average fit plot analogous
to Figure~\ref{fig:average_fits} and a NEB path between one of them
and the global minimum.

\subsection{Larger networks}

\begin{figure}[ht]
\begin{center}
\includegraphics[scale=0.5]{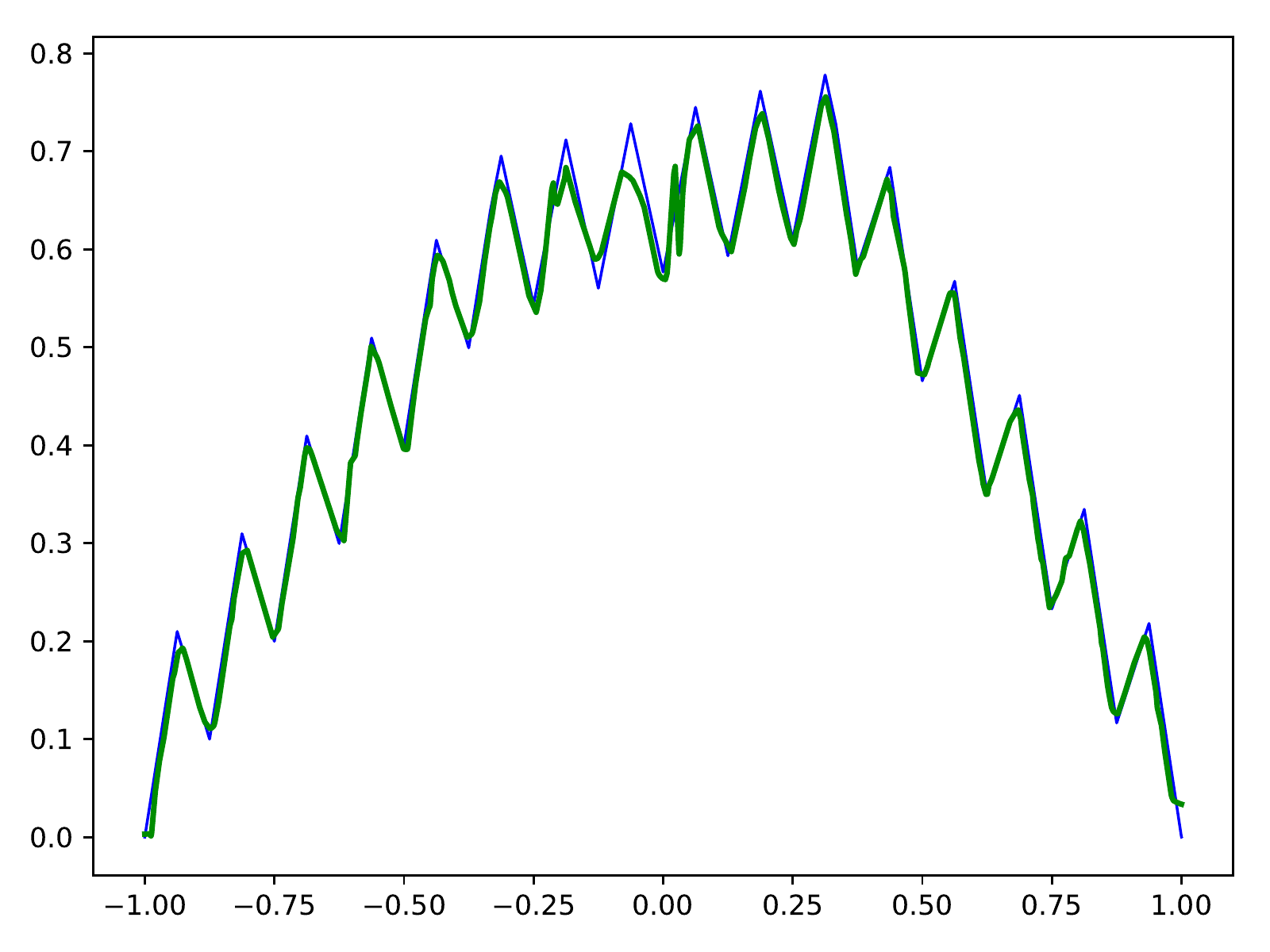}
\caption{An adadelta $15\times 15$ fit to our main $5 \times (2+3)$ example.
It is evidently quite close to a global minimum.}
\label{fig:fit_larger}
\end{center}
\end{figure}

Let $W_{K,N}$ denote the weight space of $K \times N$ networks.
Note that $W_{K,N}$ can be embedded in $W_{K',N'}$
for any $K' \ge K$ and $N' \ge N$ (and in many ways),
so it is reasonable to ask how the topography of the weight space
changes if we allow networks under training to move off the
$W_{K,N}$ submanifold and through $W_{K',N'}$.  That is,
if we train a larger model.  As we expect, this tends to turn
local minima into saddle points and allow the network
to learn more complex functions.  For example,
Figure~\ref{fig:fit_larger} shows the first adadelta fit we did
of a $10\times 10$ network to our main example function.

\section{Conclusion}
\label{sec:conclusion}

We have shown (perhaps belabored) the point that the
weight space of neural networks
can be a complicated hilly place, and there can be global minima
which are essentially unreachable using typical training techniques.
The wide, flat local minima make optimization very difficult.
In one sense, this is just the observation that nonconvex optimization
is difficult.  But this difficulty grants neural networks an interesting
opportunity: because complex, high-frequency functions are difficult to
fit, there is an implicit noise-dampening regularization built into the 
network, and training time becomes the parameter which controls how complex 
a the output of a network can be. So, in regression examples, at least,
networks of moderate size will smooth rather than reproduce the training
data.

In a perfect world, we might hope for a global-optimum oracle.
In this case, we could obtain the same sort of
regularization by limiting the effective size of our models.
This is the standard approach in non-parametric regression,
for example. Neural networks, on the other hand, with real-world
optimization techniques appear to have the rather fortuitous
property of being somewhat self-regularizing. How much we can
rely on this fact remains a model selection problem.

\end{document}